\documentclass[12pt]{article}

\usepackage{scicite}

\usepackage{times}

\usepackage{graphicx}
\usepackage{graphics}
\usepackage{color,colortbl}
\definecolor{Gray}{gray}{0.9}
\definecolor{purp}{rgb}{0.86,0.81,1}
\usepackage{comment}
\usepackage{rotating}

\usepackage{amssymb}
\usepackage{hyperref}

\usepackage{lineno}

\newenvironment{sciabstract}{%
\begin{quote} \bf}
{\end{quote}}

\title{The Future of Fundamental Science Led by Generative Closed-Loop Artificial Intelligence}

\author
{Hector Zenil,$^{1,2,3,4,\ast}$ Jesper Tegn\'er,$^{21,27}$ Felipe S. Abrah\~{a}o,$^{3,4,8,26}$, \\ Alexander Lavin,$^{19,20}$ Vipin Kumar,$^{6}$ Jeremy G. Frey,$^{7}$ Adrian Weller,$^{1,2}$ \\ Larisa Soldatova,$^{9}$ Alan R. Bundy,$^{5}$
Nicholas R. Jennings,$^{10}$ Koichi Takahashi,$^{11,12,13}$ \\Lawrence Hunter,$^{14}$ Saso Dzeroski,$^{15}$
Andrew Briggs,$^{16}$ Frederick D. Gregory,$^{17}$ \\Carla P. Gomes,$^{18}$ Jon Rowe,$^{2,22}$ James Evans,$^{23}$ \\
Hiroaki Kitano,$^{2,24}$ Ross King$^{1,2,25}$\vspace{-.5cm}\\
\noindent \\
\normalsize{$^{1}$Department of Chemical Engineering and Biotechnology, 
University of Cambridge}\\
\normalsize{$^{2}$The Alan Turing Institute}\\
\normalsize{$^{3}$Oxford Immune Algorithmics}\\
\normalsize{$^{4}$Algorithmic Nature Group, LABORES for the Natural and Digital Sciences} \\
\normalsize{$^{5}$School of Informatics at the University of Edinburgh} \\
\normalsize{$^{6}$Department of Computer Science and Engineering, University of Minnesota} \\
\normalsize{$^{7}$Department of Chemistry, University of Southampton} \\
\normalsize{$^{8}$Centre for Logic, Epistemology and the History of Science, University of Campinas, Brazil.} \\
\normalsize{$^{9}$Department of Computing, Goldsmiths, University of London} \\
\normalsize{$^{10}$Vice-Chancellor's Office, Loughborough University} \\
\normalsize{$^{11}$RIKEN Center for Biosystems Dynamics Research,} \\
\normalsize{$^{12}$RIKEN Innovation Design Office}
\normalsize{$^{13}$Keio University} \\
\normalsize{$^{14}$Center for Computational Pharmacology, School of Medicine, University of Colorado} \\
\normalsize{$^{15}$Department of Knowledge Technologies, Jozef Stefan Institute} \\
\normalsize{$^{16}$Department of Materials, University of Oxford} \\
\normalsize{$^{17}$DEVCOM ARL Army Research Office} \\
\normalsize{$^{18}$Department of Computer Science, Cornell University} \\
\normalsize{$^{19}$Pasteur Labs} 
\normalsize{$^{20}$Institute for Simulation Intelligence} \\
\normalsize{$^{21}$Living Systems Laboratory, BESE, CEMSE, King Abdullah University of Sciences and Technology}\\
\normalsize{$^{22}$School of Computer Science, University of Birmingham}
\\
\normalsize{$^{23}$Knowledge Lab, University of Chicago}
\\
\normalsize{$^{24}$The Systems Biology Institute, Okinawa Institute of Science and Technology}
\\
\normalsize{$^{25}$Chalmers Institute of Technology} \\
\normalsize{$^{26}$DEXL, National Laboratory for Scientific Computing, Brazil.} \\
\normalsize{$^{27}$Department of Medicine, Karolinska Institutet, Stockholm, Sweden.} \\
\scriptsize{$^\ast$To whom correspondence should be addressed; E-mail:  hector.zenil@cs.ox.ac.uk}
}

\date{}

\begin{document} 


\baselineskip24pt


\maketitle


\begin{sciabstract}
Recent machine learning and AI advances disrupt scientific practice, technological innovation, product development, and society. As a rule, success in classification, pattern recognition, and gaming occurs whenever there are clear performance evaluation criteria and access to extensive training data sets. Yet, AI has contributed less to fundamental science, such as discovering new principled explanatory models and equations. To set the stage for a fundamental AI4Science, we explore a perspective for an AI-driven, automated, generative, closed-loop approach to scientific discovery, including self-driven hypothesis generation and open-ended autonomous exploration of the hypothesis space. Generative AI, in general, and Large Language Models (LLMs), in particular, serve here to translate and break down high-level human or machine conjectures into smaller computable modules inserted in the automated loop. Discovering fundamental explanatory models requires causality analysis while enabling unbiased efficient search across the space of putative causal explanations. In addition, integrating AI-driven automation into the practice of science would mitigate current problems, including the replication of findings, systematic production of data, and ultimately democratisation of the scientific process. These advances promise to unleash AI's potential for searching and discovering the fundamental structure of our world beyond what human scientists have achieved or can achieve. Such a vision would push the boundaries of new fundamental science beyond automatizing current workflows and unleash new possibilities to solve some of humanity's most significant challenges.
\end{sciabstract}

\section*{Introduction}

\begin{figure}[ht!]
	\centering
	\includegraphics[width=10cm]{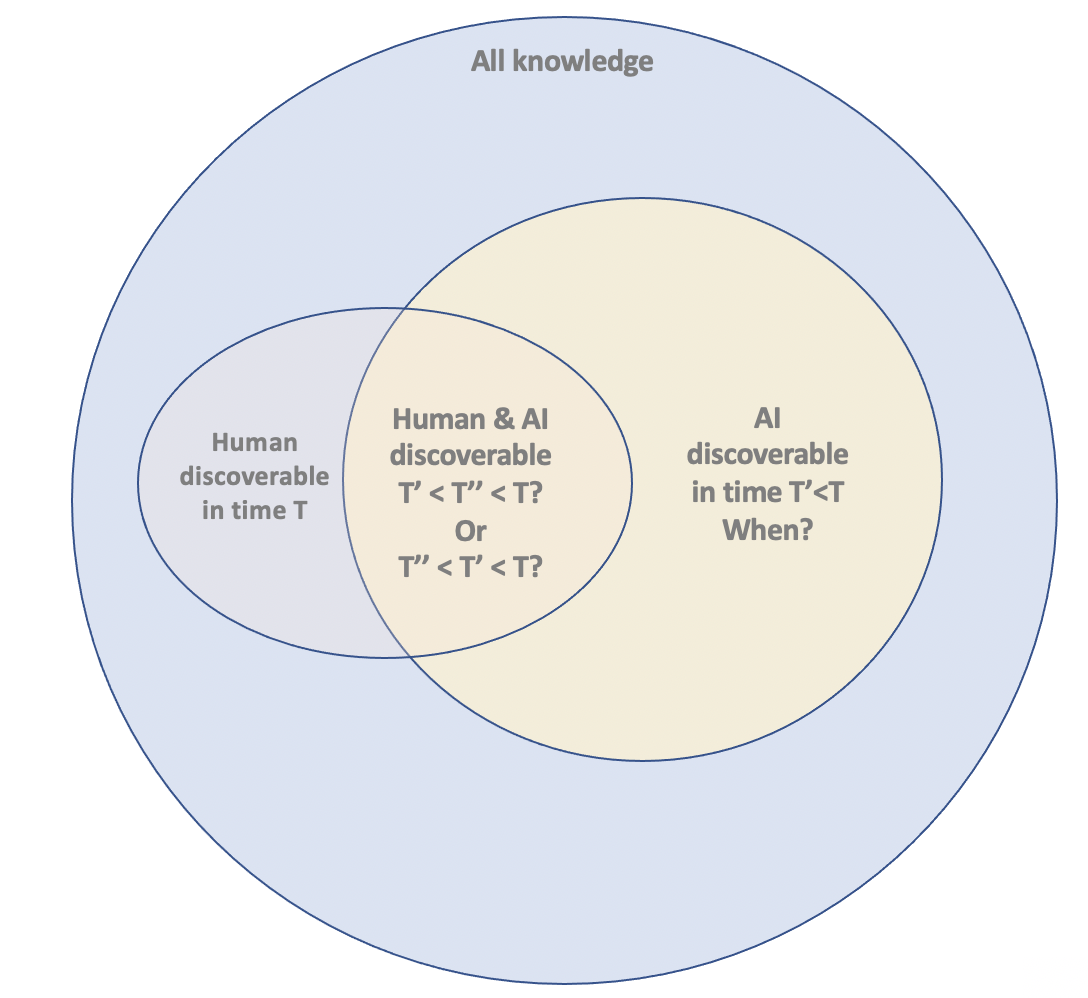}
	\caption{\label{speed}
		A quantitative framework of domain-agnostic acceleration of scientific discovery with AI, its relationship with human-carried science, and the combination of human and machine. 'All knowledge' can be interpreted as potential knowledge if it were discoverable by humans through AI or by themselves. AI time $T'$ to conduct certain tasks is traditionally taken to be faster than $T$ by orders of magnitude  \cite{King2004a} as it is also more scalable. Still, its domain-dependency and relationship to $T''$ (human-machine hybrid) are likely highly domain-specific and has traditionally ignored closed-loopness or the removal of any human input.}
\end{figure}

With the scientific revolution in the seventeenth century, the notion of mathematical modeling using equations became the efficient language of choice to understand and predict events in the natural world. Four hundred years later, we have vast amounts of data and increasing access to computational power. Recently, we have witnessed an ever-increasing comprehensive application of machine learning accelerating science in unprecedented ways with many questions around quantifying the speed up of discovery (Fig.~\ref{speed}). One consequence of this increase in scientific production that the digital revolution enabled, it is increasingly challenging for individual scientists to keep abreast of their fields and digest the relevant literature.
To advance science and perform end-to-end high-quality scientific investigations, scientists often require one or two orders of magnitude more hypothesis-led experiments than are currently humanly possible. 
Laboratories are under pressure to perform an increasing number of experiments needed to replicate results. This makes collaborations more challenging, given all the associated overheads, in particular for interdisciplinary research. It may be that some areas of science, such as new fundamental physics or theories in biology, will be too difficult for humans to advance by themselves. AI technologies may be required to be in the driving seat of knowledge discovery to continue the endeavor of human science. 

Recent advances in AI since 2010 \cite{LeCun2015,Schmidhuber2015DeeplearningNNoverview},
fuelled by large data sets and computing power, have largely targeted problems such as pattern recognition, object classification, and gaming. Yet, following this phase of difficult-to-understand and interpret - "black-box" neural network models - there has been a renewed interest in expanding the scope of machine learning models. This includes efforts to search for inspiration from the human brain 
\cite{Hassabis2017neuroscience-inspired,Zador2023Catalyzingnextgen}, to learn causality 
\cite{Pearl2009}, to incorporate different geometric priors beyond convolutional filters in the learning
\cite{Bronstein2017GeometricDLbeyond},and physics-informed machine learning
\cite{Karniadakis2021PhysicsinformedML,Lavin2021si}, to develop explanatory machine learning
\cite{Holzinger2022Explainable}. In the most recent advances moving beyond classification, there has been a growing interest in what can be referred to as AI4Science, reflected, for example, in targeted workshops at machine learning conferences such as NeurIPS and ICML 
\cite{AIforScience2023}. This is evidently a vibrant community 
\cite{Berens2023Aiforscienceemerging,Karagiorgi2022MLfundamentalphys,Raghu2020SurveyDLforsciendisc,Noe2020MLmolecularsim,Richards2019DLneuro} active in numerous scientific areas. Naturally, much of the ``early" focus has been on data integration, refining measurements, augmenting data, optimising parameters, automation of workflows and data analysis, and scientific visualisation 
\cite{Wang2023ScientificAI}. Most recently, with the emergence of foundational models 
\cite{Bommasani2022OpportunitiesRisksmodels}, based on the transformer architecture for large language models 
\cite{OpenaiGPT42023}, there is the idea that given large enough data, we can train foundational models which may learn emergent properties to ``understand'' their respective domains of knowledge
\cite{Srivastava2023beyondImitation}. This is, however, an open question and challenge for the field. Do we need intelligent priors, or is a huge amount of data sufficient? Yet, regardless of the resolution of this current debate, there is, in our view, a distinct gap between all the exciting recent progress versus having systems, corresponding to an artificial scientist. Such an agent would truly discover and formulate new scientific laws using fundamental mathematical-physical models from observations. Here, in this perspective, we put forward a formulation of a closed-loop iterative formulation as a practical path towards this end.

To further conceptualise how AI can augment science, we like to distinguish between the following levels. First, AI and machine learning can operate as extractors of information. This includes text mining of scientific literature to find relevant papers and, in the best case, extract knowledge and synthesise a body of research from vast sources. A more ``modern" use of AI, as mentioned above, has made existing workflows or procedures more efficient, such as being faster and more automatic. This includes augmented computations and simulations in physics. Alternatively, to make an analysis workflow more automatic by constructing a loss function that incorporates several parameter-dependent steps into a single (complex) optimisation problem. The first version of AlphaFold is one example
\cite{Senior2020Improved}. However, at these two levels, AI primarily supports and enhances current scientific practice. A third level, which is the target in the present perspective,  is where AI could potentially discover and learn novel scientific laws, thus finding a ``true'' representation of a process in nature. For example, a useful compressed latent representation could be learned by training an AI system on data. Alternatively, the scientist could impose soft priors such that certain symmetries and invariants exist in the problem, thus forcing the AI system to discover interpretable structures in the physical process. Geometric machine learning is similar to the ``classical" model-based analysis of nature initiated by the scientific revolution. Yet, it is an open problem, how to find such useful and "true" priors from observations in contrast to impose them as regularizing conditions as priors.  Here in this review, we focus on the prospect of not only augmenting science, but also by finding such useful, interpretable representations leading to new scientific discoveries by involving AI in what we refer to as a closed-loop-science-AI iterative cycle. 

AI can speed up the scientific discovery process and has the potential to advance AI itself in areas relevant to fundamental science, such as causal discovery, automation of experimental science, and the expansion of scientific knowledge.  One current bottleneck is knowledge representation, which, by nature, is biased toward the limited understanding and application of the human (scientific) endeavour.  A closed-loop-science-AI iterative cycle would be integrated with laboratory automation to execute cycles of planned experiments \cite{King2004a,King2009}. However, the acceleration from automating and closing the loop of the whole scientific cycle leaves unanswered questions, some of which are illustrated in Fig.~\ref{speed}. These systems can fully automate simple forms of scientific research
and can further facilitate collaboration across disciplines and between partners--humans or AI systems, but the actual performance of each combination is still unknown and it is still possible that human-machine collaboration produces the most productive outcome and the most relevant to human science too, as it may remain guardrailed to purely human interests. This, however, may also deprive us from undertaking less human explorations that may have led to results of human interest. Another key question is whether teaming up or AI alone would allow us to cover a larger region of `human knowledge' defined recursively as the knowledge that can potentially be reached by human understanding.


In contrast to finding new laws or representations, the applications of AI that have been successful so far have been largely limited to industrial applications, classification problems, and data science.A recipe for success has been to pick a problem that has a well-defined performance metric \cite{McCarthy2007-ns}. Problems should preferentially have a history of previously increasingly successful attempts to solve them. Examples include board games (Chess, GO) and bioinformatics workflows (transcription factor binding, protein folding, antibiotics)
\cite{Silver2018GeneralReiforLearn,Stokes2020DLantibiotic,Alipanahi2015PredictingDNARNADeepLearn,Senior2020ImprovedproteinDL}. 
Yet, despite being impressive, the success of these examples, hinges upon clever search algorithms and efficient implementation of end-to-end workflows. But, in the end, no new fundamental laws of nature are being discovered. Furthermore, as a reflection of the lack of fundamental laws, the inner workings of these AI systems remain challenging to explain and disentangle. To advance beyond this state of affairs, we argue that we need AI systems that can discover new transparent representations and generative laws of the scientific problem at hand. The notion of an iterative closed-loop discovery scheme constitutes one putative path forward. Yet, only a few successful examples have closed the full loop of scientific discovery \cite{ROSS}. 

Human scientists today need to think about how to create AI systems that can partner with scientists and take on responsibilities over the complete arc of scientific discovery \cite{Wang2019}: from the process of observation and intervention to hypothesis generation; from a domain knowledge base to conducting experiments and evaluating results; and from rejecting or validating the assumptions to integrating them into the current knowledge base and filing them with the relevant existing literature.
Thus, the question is how to make substantial and meaningful advances in AI to enable us to go even further in accelerating science, hitherto driven exclusively by humans,
to not only rapidly expand human knowledge and improve the impact of scientific practice, but also to increase its reliability, availability, reproducibility, verifiability, transparency, and trustworthiness as the processes involved in scientific discovery become more automated.

In Fig.~\ref{speed}, we propose some quantitative measures that will not apply to all cases but rather instances where a combination of AI and human approaches can further accelerate science. Nevertheless, the expectation is that AI will provide a real gain on most fronts and domains.

\section*{AI in Scientific Discovery}\label{sectionPreviouswork}

\subsection*{\textit{Challenges}}


Humans are traditionally biased and prone to very well-known cognitive fallacies or biases, which science is hardly a stranger to~\cite{Nosek2012,Fanelli2017,Nuzzo2015}. One common and increasingly discussed issue is reproducibility\index{reproducibility} across all domains~\cite{Goodman2018,Harris2019}. 
Humans are ill-equipped to deal with the repetitive tasks that reproducibility entails, and there are all sorts of inducements for consciously or unconsciously making dubious moves, particularly when it comes to the game of funding and high-impact publishing~\cite{Kaanders2021}. 
Confirmation bias, fake rigour, prior assumptions/hypotheses omission, ad hoc methodologies, cherry-picking experimentation, selective data, hype and overstatement of results, network community effects, ``rich-get-richer'' phenomena widening the inequality gap in science, and poor reporting are examples~\cite{Fanelli2017,Nuzzo2015,BarabSci,Fortunato2018,Nature2016,Colizza2006-ed,Baker2016-cd}.

We used to think that science was entirely objective, but history has taught us that it is also driven by community choices and groups, where it becomes clear that political and social preferences and underlying cognitive biases can interfere with scientific progress~\cite{Baddeley2015-jp,Resnik2016,Evans2011}.
All these problems are leading to a crisis impacting scientific networks, putting collaborative networks at a disadvantage and favouring competitive ones, and often compromising the very principles of scientific practice.

Closed-loop-AI-led science has the potential to mitigate all these problems because it can bootstrap itself with the right mechanisms to detach itself from human-led science and its own biases, even if human scientists initially transfer them. Furthermore, this invites scientists with the task of initially guiding AI as to the type of meaningful research that should be conducted but then letting it explore regions of the scientific space that may never be reachable by human scientists while having the option to keep what human scientists believe is of greatest interest but letting the close-loop-AI system to potentially continue using less human-relevant content searching for novelty in terms of what is potentially interesting to go after. That is to have AI bootstrap itself out of and above the loop-AI-science without human guidance.

One challenge in this direction is that automation can easily fall into the over-fitting trap without human input, and mechanisms to avoid this must be in place. However, it has been found that simplicity and randomness are powerful mechanisms to avoid local minima and maxima when iterating over searching algorithms\cite{HernandezOrozco2021}. 
A striking feature of supervised machine learning is its propensity for over-parametrisation~\cite{Venturi2019-mr}. Deep networks contain millions of parameters, often exceeding the number of data points by orders of magnitude, so often, the model starts to over-fit right at the beginning~\cite{Goodfellow2016}. 
Broadly speaking, networks are designed to interpolate the data, learning/constructing an associated manifold by driving the training error to zero. 
Deep neural networks\index{Deep neural networks} in particular are widely regarded as black-box approaches, ill-equipped to offer explanations of the produced models for classification, often with superhuman ability~\cite{blackbox, Rudin2019-pv}. One strategy that has enabled researchers to make progress in understanding the workings and limitations of deep learning is the use of what has been called  `generative models'~\cite{Salakhutdinov2015-su}.\index{generative models} This involves training adversarial algorithms represented by neural networks that systematically tamper with data while asking it to generate novel examples~\cite{Creswell2018-qa, Bian2021-vh}. By observing the resulting examples and how the classifier fails, they can understand the model's limitations and improve the classifier.

\begin{figure}[ht!]
	\centering
	\includegraphics[width=12cm]{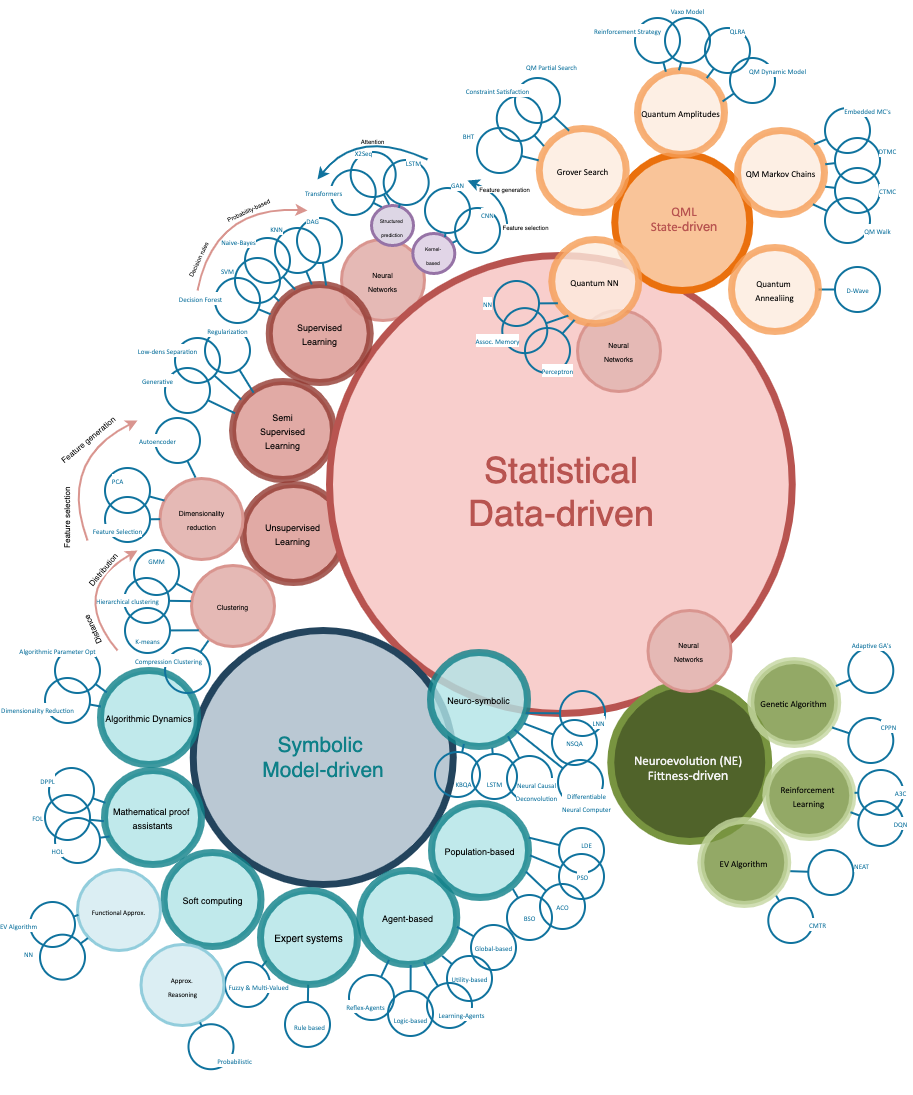}
	\caption{\label{aistate}
		Bubble landscape of current approaches to AI from and for science. Bubbles may occur more than once when related to several larger domains. Some approaches may have alternative names or have been re-branded in certain contexts. Neuro-symbolic models have sometimes been referred to as 'intuitive' while some statistical-driven approaches have been labelled as `cognitive computing'. Generative AI (GenAI) has made little to no contributions to fundamental science so far but has great potential. Large Language Models (LLMs) may significantly tap into and contribute to the exploratory capabilities of the scientific hypothesis space, given their capabilities to process human language in which all human science has been written. GenAI and LLMs are approaches of statistical nature, but it remains unexplored to what extent they may develop symbolic capabilities from statistical (e.g. linguistic) patterns.}
\end{figure}

However, current approaches in science (see Fig.~\ref{aistate}), including most machine and deep learning methods, rely heavily on traditional statistics and information theory. Consequently, such models are insufficient to capture certain fundamental properties of data and the world related to recursive and computable phenomena, and they are ill-equipped to deal with high-level functions such as inference, abstraction, modelling, and causation, being fragile and easily deceived \cite{Zenil2020,Scholkopf2021,Colbrook2022}, for example because they are prone to finding spurious patterns in large data sets \cite{Calude2017,Smith2020}.

Most of these algorithms fail to be scalable in domains outside the training set. Such algorithms lack mechanisms for abstraction and logical inference, they fail at generalisation~\cite{Nadeau2003-vk}. For example, in the case of driverless cars\index{driverless cars}, one does not want a car to crash millions of times to learn how not to crash, so current techniques such as adversarial networks offer a way to produce examples in which not driving appropriately can lead to an event that is labelled a crash~\cite{Spooner2021-px}. However, driving and crashing are events where cause and effect need to be learned, which current approaches cannot do.



When AI leads science so that laboratory experiments are automated to execute cycles of planned experiments, AI frees humans from repetitive, tedious, and error-prone tasks and can deal with vast amounts of data that no human could handle \cite{Kitano2016}. 
These human scientists, in turn, can feed the AI systems back with new insights and novel theories.
Thus, such an emerging feedback loop of AI-human collaboration will synergistically boost scientific discovery toward previously unattainable results, rigour, and dissemination.

To overcome the above limitations and challenges, we claim that it will require the fostering of new theories and methods, as well as human and technological resources in AI, data science, and interdisciplinarity, so scientists become capable of dealing with this AI-human interplay both at an infrastructural and a metastructural level.
One of these theories may involve developing mathematical frameworks that can deal with the fact that not only the empirical findings, but also the new theories themselves the scientists within the loop are devising can be influenced by other AI algorithms within the loop, and vice versa.
For example, this may require causal analysis (or inverse-problem solving) \cite{Zenil2019b} when both the observer and the observed system are mutually perturbing the underlying generative model of each other \cite{Zenil2020cnat,Abrahao2022}.
One of these methods may involve AI that guides AI, and translates results to humans, and this intermediate AI may not be of the same type.  For example, causal and model-driven AI \cite{Jin2021-jw,Thieme2005-ij} may be required to disentangle other AI systems to which human scientists cannot relate if they do not have a mechanistic explicative component, whether there is one or not. This may lead to some sort of meta-AI capable of dealing with knowledge representations at a meta-level \cite{Evans2011}, which includes the network dynamics of the each agent (whether AI or human) in the loop, so that this meta-AI still remains explainable to humans \cite{Goebel2018-cq}. 
This may not require Artificial General Intelligence but would require a different set of skills than purely statistical machine learning approaches.


\subsection*{\textit{Historical Context}}

Applications of AI in science are quite broad and cover many fields. The idea of automating reasoning goes back to Leibniz, where the modern incarnation can be traced back to efforts to build computing machines in Europe. In particular, the heroic efforts of Alan Turing's\index{Turing, Alan} work at Bletchley to automate the problem of code breaking and his ideas of an imitation game \cite{Turing1,Turing2}. It can also be traced back to Joshua Lederberg\index{Lederberg, Joshua} (Nobel laureate) \cite{Lederberg}, Ed. Feigenbaum\index{Feigenbaum, Edward} (Turing award winner) \cite{Feigenbaum1}, Karl Djerassi\index{Djerassi, Karl} (co-inventor of the contraceptive pill) \cite{Djerassi}, and colleagues at Stanford in the 1960s, who worked on automating mass-spectroscopy for the Viking Mars lander \cite{DENDRAL,Buchanan1984}. AI has long been a tradition of taking scientific discovery as an area of study. In the 1970s the Nobel Prize laureate and Turing prize winner Herbert Simon\index{Simon, Herbert} developed Bacon, an AI system for science \cite{Langley1987}. Since this pioneering work, much has been achieved, and there are now many convincing examples of AI systems making clear contributions to scientific knowledge (e.g. the very recent \cite{Burger2020,Jumper2021-gb}).  

Eurisko\index{Eurisko} \cite{lenat} and Cyrano\index{Cyrano} \cite{hasse} are two examples of other attempts to perform automated discovery from basic principles in a variety of technical fields, in particular in mathematics, chemistry, and a few other domains. 
These are systems that can be viewed as heuristic search systems, with the additional advantage that they can reconfigure their own search space.

Some commercial products are specifically designed to be applied to knowledge and scientific discovery. For example, DataRobot \cite{DataRobot} \index{DataRobot} promotes Eureqa \cite{Eureqa},\index{Eureqa} having acquired Nutonian \cite{Nutonian,Dubcakova2011-he, Awange2018-el}.\index{Nutonian} Eureqa was designed to create models from time series data and is based on creating random equations from mathematical building blocks through evolutionary search to explain the data~\cite{Dubcakova2011-he}. It has been called a ``Virtual Data Scientist''\index{Virtual Data Scientist} \cite{Eureqa}. 

A team of researchers from Google DeepMind\index{DeepMind} launched a machine learning project called AlphaFold\index{AlphaFold} in 2018 to participate in the Critical Assessment of Techniques for Protein Structure Prediction or CASP~\cite{Wei2019-vt}. CASP\index{CASP} is a biennial competition that assesses state-of-the-art three-dimensional protein structure modelling. In its first version, AlphaFold\index{AlphaFold}  was particularly successful at predicting the most accurate structure for targets rated as the most difficult by the competition's organisers, but it was not until the second program, AlphaFold 2,\index{AlphaFold 2} in 2020, when the team achieved a level of accuracy much higher than any other group before and scored above 90 for around two-thirds of the proteins in CASP's global distance test (GDT), a test that measures the degree to which a structure predicted by a computational program is similar to the structure validated experimentally, with 100 being a complete match. AlphaFold\index{AlphaFold} relied on a lot of human knowledge already generated in the years before, especially in areas such as molecular dynamics. The program was designed to include the expert domain in the form of the training data. How much molecular biological knowledge was introduced is still not known, but while it required a team that did draw heavily on domain expertise to tune it, most of the predictive power came from the AlphaFold 2\index{AlphaFold 2} tool itself~\cite{Jumper2021-gb,Skolnick2021-ty}.

A precursor of AI in physics is the project GALILEO (Guided Analysis of Logical Inconsistencies Leads to Evolved Ontologies)~\cite{Liu2021-zj}. The GALILEO project tried to model the repair of faulty theories of Physics whose predictions were contradicted by empirical evidence. 
One area of successful application of machine learning from climate data, for example, was the discovery of climate dipoles through machine learning~\cite{Liu2021-zj}. 
Physics-driven AI has the potential to impact how we approach science, on our current predominantly data-reliant---as opposed to the model-centred---scientific method, by placing the mechanistic model at the centre of modelling itself.
Paradoxically, current physics-led AI and machine learning research have distracted researchers from more fundamental research, even though the discussion has started, and researchers will hopefully eventually get around to the first principles they claim to care about.

On the knowledge side, there are many applications of knowledge extraction of interest, such as for drug re-purposing by pharmaceutical companies~\cite{Gupta2021-py, Liu2021-repurpose}. 
On task-oriented problem solving, we can find an increasing number of workflow systems that understand scientific tasks and carry them out.
There have been some success stories demonstrating that by collecting and integrating available molecular data into computational models, accurate predictions of interventions in the system can actually be made. An example is the Robot Scientist program \cite{King2004a} that was able to autonomously execute high-throughput hypothesis-led 
research investigating yeast-based functional genomics, with the next-generation scientific program later using the same principles for drug screening. In another example, a computational model of Halobacterium salinarum NRC-1 was first constructed through massive data integration and machine learning-driven inference of the regulatory network \cite{Bonneau2007}. 

Another example was the ambitious whole-cell computational model of the life cycle of the human pathogen Mycoplasma genitalium \cite{Karr2012}. The model accounted for all annotated gene functions and was validated against a broad range of data. Now, the model encompasses approximately 500 genes and their interactions.

In the area of neural networks, there has been, for example, an effort to make them `understand' cause and effect by algorithmic training. While more research is needed, fundamental research is aware that alternative approaches are required to capture the complexities of hypothesis and model generation or selection~\cite{Luo2020-xz,Scholkopf2021,HernandezOrozco2021}. 
In this sense, the research in this type of higher-order AI, such as deconvolution from searching for generative processes from the entire algorithmic space \cite{Zenil2019b}, will also be crucial to advance current research. 	

To present a summary of the current state of AI applications to each scientific domain, Table~\ref{tab:motifs} 
displays an organisation of scientific domains\footnote{Note that: 
\textit{Complexity} includes systems and intelligence as defined by the Santa Fe Institute;
\textit{Manufacturing} notably includes ML-based design of sensors and chips; 
and \textit{Earth systems} includes oceans, land, air, and near space (see \href{https://earthdna.org}{earthdna.org}).} 
and the applicable AI algorithms' classes and approaches. Scientific domains are approximately ordered from smallest physical scales to largest.
Overlapping areas are not reflected in this high-level table (e.g., semi-supervised RL methods, or the representation of neural networks (NNs) that conflates various deep learning types like LSTM and Transformers), not to mention complex, context-dependent multidisciplinarity. Table~\ref{tab:motifs}'s content was the consensus and understanding of a subset of this paper authors. While supervised statistical methods have contributed to almost every area of knowledge, these are of very different type mostly ranging from identification to classification. Some areas are more difficult than others across all approaches, such as mathematics, philosophy, and epistemology.  In general, statistical approaches rank poorly at finding first principles or adding new mechanistic knowledge to scientific domains.

Generative AI (GenAI) and Large Language Models (LLMs) are promising to advance science by assimilating and synthesising the vast corpus of human knowledge embedded in scientific literature. Through this synthesis, LLMs can interconnect disparate ideas, construct unique hypotheses, and venture into uncharted areas of scientific knowledge. However, this exploration is bound by the data they have been trained on, creating a theoretical bubble that could lead to model collapse through excessive training on the same data.

To burst this bubble, it is essential to supplement LLMs with other methods and multiple sources. For instance, active learning could serve to maximise information gain, challenging the model with fresh data and different viewpoints cross-pollinating from different scientific domains. Hybrid models blending AI with symbolic reasoning could tackle scientific problems requiring high-level abstraction, thus broadening LLMs' capabilities. This approach would therefore fall into the neuro-symbolic category for purposes of scientific discovery.

Indeed, an area where LLMs could be especially impactful is in scientific model discovery. By analysing patterns and correlations in vast datasets, LLMs could help identify mathematical relations and possibly reveal new potential (physical, or computational) laws just as it learns language grammar from natural language statistics. This could expedite the scientific process, enabling more rapid breakthroughs.

Furthermore, LLMs could make a significant contribution to causal analysis. By processing extensive scientific literature, they could draw links between causes and effects that might be overlooked by human researchers, proposing novel causal hypotheses for testing. Pairing this with counterfactual reasoning, where the AI predicts the outcome of modifying specific variables, could deepen our understanding of cause-effect relationships, and help simulate alternative model outcomes.

However, in addition to inheriting the limitations from statistical machine learning in general \cite{Colbrook2022,Abrahao2021darxiv}, it is also important to acknowledge the limitations of current LLMs.
They currently lack the depth needed for any breakthrough to happen and require quality and diversity of data allowing an LLM `temperature' (favouring less likely statistical patterns) to explore further along the potential long tails of the distribution of scientific results with potential breakthrough science away from incremental average science. A collaborative approach, in which human scientists guide the AI, can help harness the strengths of both worlds, mitigating the current weaknesses of LLMs and statistical ML, ensuring more effective utilisation of this technology today. 

\begin{table}
	\centering
	\vskip 7em
	\hskip -3em
	\footnotesize
	\begin{minipage}{\linewidth}
		\begin{tabular}{l|cccccccccccccccccccccc}
			& \begin{rotate}{90} Supervised - NN and kernel-based \end{rotate}
			& \begin{rotate}{90} Supervised - probabilistic and rules-based \end{rotate}
			& \begin{rotate}{90} Semi-supervised \end{rotate}
			& \begin{rotate}{90} Generative \end{rotate}
			& \begin{rotate}{90} Unsupervised \end{rotate}
			& \begin{rotate}{90} Algorithmic dynamics \end{rotate}
			& \begin{rotate}{90} Soft computing and probabilistic numerics \end{rotate}
			& \begin{rotate}{90} Expert systems \end{rotate}
			& \begin{rotate}{90} Model-based  \end{rotate}
			& \begin{rotate}{90} Structural causal \end{rotate}
			& \begin{rotate}{90} Neurosymbolic \end{rotate}
			& \begin{rotate}{90} Reinforcement Learning \end{rotate}
			& \begin{rotate}{90} Neuroevolution \end{rotate}
			& \begin{rotate}{90} Genetic Algorithm \end{rotate}
			& \begin{rotate}{90} Open-ended search, evolution \end{rotate}
			& \begin{rotate}{90} Quantum Amplitudes \& Grover search \end{rotate}
			& \begin{rotate}{90} QM Markov Chains \& Simulation \end{rotate}
			& \begin{rotate}{90} Data-driven, supervised Quantum (NNs) \end{rotate}
			\\
			\hline
			{Mathematics}& - & $\sim$ & - & $\sim$ & - & \checkmark & - & \checkmark & - & - & $\sim$ & $\sim$ & - & - & - & - & - & - \\
			
			\rowcolor{purp}
			{HE Physics - theo}& - & $\sim$ & - & $\sim$ & - & $\sim$ & - & \checkmark & - & $\sim$ & $\sim$ & - & - & - & $\sim$ & $\sim$ & \checkmark & - \\
			
			\rowcolor{purp}
			{HE Physics - exp}& \checkmark & \checkmark & \checkmark & \checkmark & - & - & \checkmark & - & \checkmark & $\sim$ & $\sim$ & \checkmark & $\sim$ & - & - & \checkmark & \checkmark & \checkmark \\
			
			{Optics \& Acoustics}& \checkmark & \checkmark & $\sim$ & \checkmark & $\sim$ & $\sim$ & \checkmark & - & $\sim$ & - & $\sim$ & $\sim$ & - & $\sim$ & - & $\sim$ & - & - \\
			
			{Complexity}& - & - & $\sim$ & \checkmark & $\sim$ & \checkmark & - & \checkmark & \checkmark & \checkmark & $\sim$ & \checkmark & $\sim$ & \checkmark & \checkmark & - & $\sim$ & - \\
			
			{SynBio \& Ind Biotech}& \checkmark & \checkmark & $\sim$ & \checkmark & \checkmark & $\sim$ & $\sim$ & \checkmark & \checkmark & $\sim$ & $\sim$ & \checkmark & $\sim$ & \checkmark & $\sim$ & - & $\sim$ & $\sim$ \\
			
			\rowcolor{purp}
			{Organic Chemistry}& \checkmark & \checkmark & \checkmark & \checkmark & \checkmark & \checkmark & \checkmark & \checkmark & \checkmark & \checkmark & \checkmark & \checkmark & \checkmark & \checkmark & \checkmark & \checkmark & \checkmark & \checkmark \\
			
			\rowcolor{purp}
			{Physical Chemistry}& \checkmark & \checkmark & \checkmark & - & - & - & - & - & - & - & - & - & - & - & - & - & - & - \\
			
			\rowcolor{purp}
			{Electrochemistry}& \checkmark & \checkmark & - & - & - & - & - & - & - & - & - & - & - & - & - & - & - & - \\
			
			{Materials}& \checkmark & \checkmark & $\sim$ & \checkmark & $\sim$ & - & - & \checkmark & - & $\sim$ & $\sim$ & \checkmark & $\sim$ & $\sim$ & $\sim$ & - & $\sim$ & $\sim$ \\
			
			{Computing}& \checkmark & \checkmark & - & \checkmark & - & - & $\sim$ & - & - & - & $\sim$ & \checkmark & $\sim$ & \checkmark & - & $\sim$ & \checkmark & \checkmark \\
			
			\rowcolor{purp}
			{Medicine, molecules/proteins}& \checkmark & \checkmark & - & $\sim$ & - & $\sim$ & - & $\sim$ & $\sim$ & $\sim$ & $\sim$ & \checkmark & $\sim$ & \checkmark & $\sim$ & - & - & $\sim$ \\
			
			\rowcolor{purp}
			{Medicine, drug development}& \checkmark & \checkmark & $\sim$ & $\sim$ & - & \checkmark & $\sim$ & \checkmark & \checkmark & \checkmark & \checkmark & \checkmark & $\sim$ & $\sim$ & - & - & - & - \\
			
			\rowcolor{purp}
			{Medicine, clinical}& \checkmark & \checkmark & - & \checkmark & - & $\sim$ & $\sim$ & \checkmark & \checkmark & \checkmark & $\sim$ & $\sim$ & - & - & - & - & - & - \\
			
			{Botany \& Zoology}& \checkmark & \checkmark & $\sim$ & $\sim$ & $\sim$ & $\sim$ & - & - & - & $\sim$ & - & - & - & \checkmark & \checkmark & - & - & - \\
			
			{Systems bio \& epidemiology}& $\sim$ & \checkmark & $\sim$ & \checkmark & - & $\sim$ & \checkmark & \checkmark & \checkmark & \checkmark & \checkmark & \checkmark & $\sim$ & \checkmark & \checkmark & - & $\sim$ & - \\
			
			{Neuro and Cog sciences}& \checkmark & \checkmark & $\sim$ & \checkmark & $\sim$ & $\sim$ & \checkmark & - & \checkmark & $\sim$ & \checkmark & \checkmark & $\sim$ & $\sim$ & $\sim$ & $\sim$ & \checkmark & $\sim$ \\
			
			\rowcolor{purp}
			{Energy - nuclear (fis/fus)ion }& \checkmark & \checkmark & - & $\sim$ & - & $\sim$ & $\sim$ & \checkmark & \checkmark & $\sim$ & $\sim$ & \checkmark & $\sim$ & - & $\sim$ & $\sim$ & $\sim$ & $\sim$ \\
			\rowcolor{purp}
			{Energy, generation \& storage}& \checkmark & \checkmark & - & \checkmark & - & $\sim$ & $\sim$ & \checkmark & \checkmark & $\sim$ & - & \checkmark & $\sim$ & \checkmark & $\sim$ & $\sim$ & - & $\sim$ \\
			
			\rowcolor{purp}
			{Energy, oil \& gas}& \checkmark & \checkmark & $\sim$ & - & - & - & - & \checkmark & $\sim$ & $\sim$ & - & \checkmark & - & - & $\sim$ & - & - & - \\
			
			{Manufacturing}& \checkmark & \checkmark & - & - & $\sim$ & - & - & \checkmark & - & - & $\sim$ & $\sim$ & - & $\sim$ & - & - & - & - \\
			
			{Engineering \& Industrials}& \checkmark & \checkmark & - & $\sim$ & \checkmark & - & $\sim$ & \checkmark & \checkmark & $\sim$ & $\sim$ & \checkmark & - & - & $\sim$ & - & - & $\sim$ \\
			
			\rowcolor{purp}
			{Energy Systems}& \checkmark & \checkmark & - & $\sim$ & \checkmark & - & \checkmark & \checkmark & \checkmark & $\sim$ & $\sim$ & $\sim$ & - & - & $\sim$ & - & - & - \\
			
			\rowcolor{purp}
			{Transp. \& Infrastructure}& $\sim$ & \checkmark & - & - & $\sim$ & - & - & \checkmark & \checkmark & $\sim$ & - & $\sim$ & - & - & $\sim$ & - & - & - \\
			
			{Agriculture}& $\sim$ & \checkmark & $\sim$ & \checkmark & $\sim$ & - & $\sim$ & \checkmark & \checkmark & \checkmark & $\sim$ & \checkmark & - & $\sim$ & $\sim$ & - & - & - \\
			
			
			{Ecology}& \checkmark & \checkmark & $\sim$ & \checkmark & $\sim$ & - & $\sim$ & \checkmark & \checkmark & \checkmark & $\sim$ & $\sim$ & $\sim$ & \checkmark & \checkmark & - & - & - \\
			
			\rowcolor{purp}
			{Socioeconomics \& Markets}& \checkmark & \checkmark & $\sim$ & - & - & $\sim$ & $\sim$ & \checkmark & \checkmark & $\sim$ & $\sim$ & \checkmark & $\sim$ & \checkmark & $\sim$ & - & - & - \\
			
			\rowcolor{purp}
			{Finance}& \checkmark & \checkmark & - & - & - & - & - & \checkmark & \checkmark & $\sim$ & $\sim$ & \checkmark & - & \checkmark & - & - & - & - \\
			
			\rowcolor{purp}
			{Politics \& Geopolitics}& - & $\sim$ & - & $\sim$ & $\sim$ & - & $\sim$ & \checkmark & $\sim$ & $\sim$ & - & $\sim$ & - & $\sim$ & $\sim$ & - & - & - \\
			
			{Defense, aerospace}& \checkmark & \checkmark & \checkmark & - & - & \checkmark & \checkmark & \checkmark & $\sim$ & $\sim$ & $\sim$ & $\sim$ & $\sim$ & $\sim$ & - & - & - & - \\
			
			\rowcolor{purp}
			{Climate, weather}& $\sim$ & \checkmark & \checkmark & \checkmark & $\sim$ & - & - & $\sim$ & \checkmark & $\sim$ & \checkmark & $\sim$ & $\sim$ & \checkmark & \checkmark & - & $\sim$ & $\sim$ \\
			
			\rowcolor{purp}
			{Earth Systems}& $\sim$ & \checkmark & $\sim$ & $\sim$ & $\sim$ & $\sim$ & $\sim$ & $\sim$ & \checkmark & $\sim$ & \checkmark & $\sim$ & $\sim$ & $\sim$ & $\sim$ & - & $\sim$ & - \\
			
			{Astrophysics \& Cosmology}& \checkmark & \checkmark & \checkmark & \checkmark & \checkmark & \checkmark & - & - & $\sim$ & $\sim$ & - & - & $\sim$ & $\sim$ & $\sim$ & $\sim$ & $\sim$ & $\sim$ \\
			
			{Philosophy, Epistemology}& - & - & - & - & - & \checkmark & - & $\sim$ & - & $\sim$ & - & - & - & $\sim$ & \checkmark & - & - & - \\
			
			\hline
		\end{tabular}
	\end{minipage}
	\vskip 0.3em
	\caption{
		Scientific domains  and the applicable AI algorithms' classes and approaches. '-' and \checkmark means simply no (or unknown) and yes, respectively; 
		$\sim$ implies the ML application is likely but not yet done nor sufficiently validated. This table is very dynamic and requires an update every quarter given the speed of developments impossible to keep up with without considerable effort or help from AI.}
	\label{tab:motifs}
\end{table} 

\newpage

\section*{Aspects of AI-Led Closed-Loop Science}\label{sectionDiscoverycycle}

The ability to predict and design (inverse design), while exceptionally useful, will not necessarily lead to new fundamental discoveries (new theories) unless AI and human goals in scientific discovery are aligned and synergistically intertwined to impose similar objectives quantified and introduced, for example,  a loss function .

\begin{figure}[ht!]
	\centering
	\includegraphics[width=16cm]{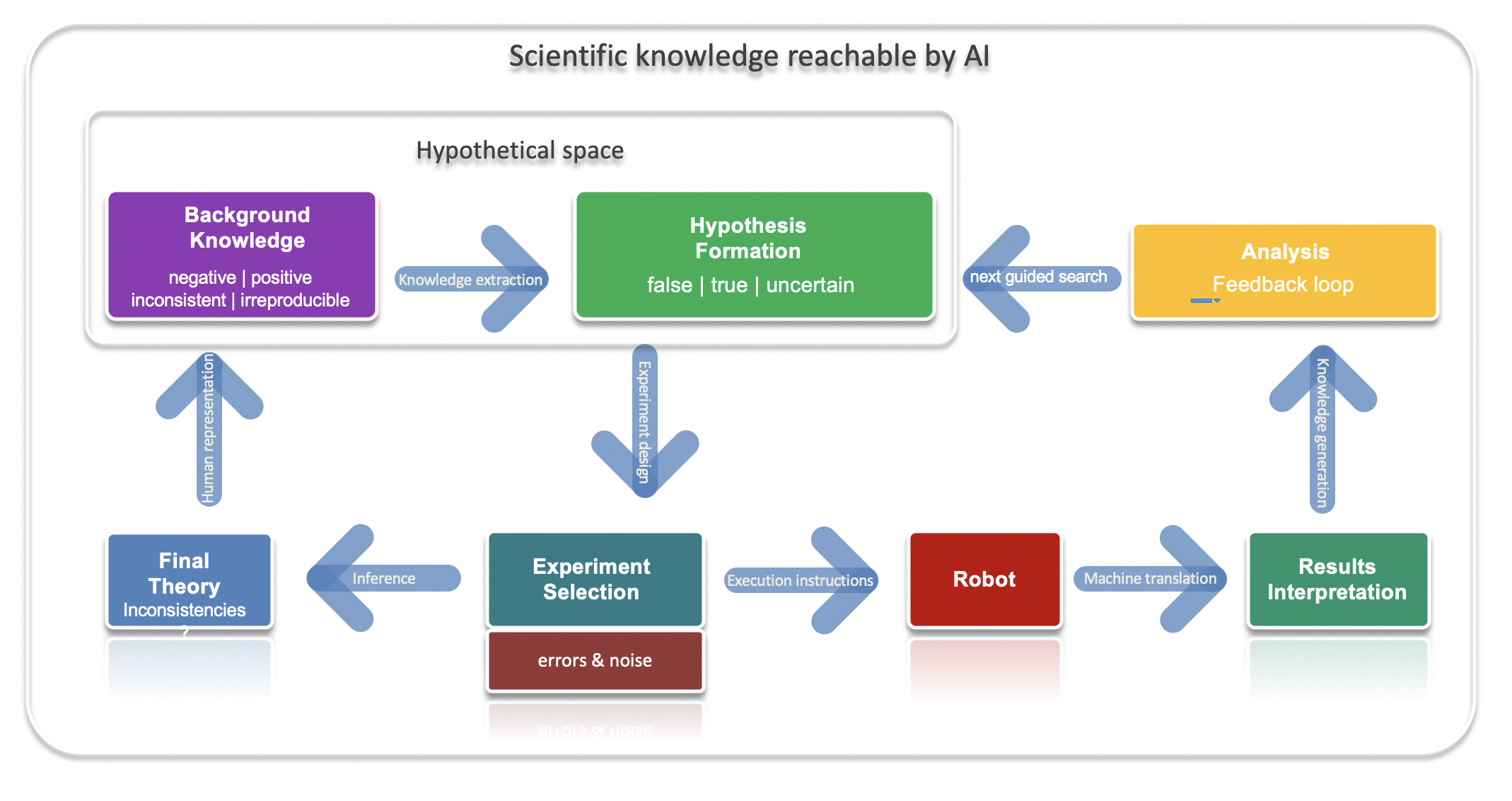}
	\caption{\label{closeloop}Visual representation of closed-loop full experimentation cycle for scientific discovery pathways, adapted and combining ideas from~\protect\cite{Kitano2016} and~\protect\cite{King2004a}. LLMs can now facilitate closing this loop but require help to connect each module and process in a causal rather than only a statistical fashion.}
\end{figure}

This is because scientific discovery cycles, such as those illustrated in Figs.~\ref{closeloop}, are not isolated parts but belong within a greater cycle of scientific inquiry spanning an entire topic or field comprised of a community of scientists.

It is the larger learning cycle that fuels the questions in the smaller learning cycles. 
The larger cycle is fuelled by human curiosity and human challenges and has a strong historical and social component, but the shorter cycles, being more well-defined, they are more prone to be automated. 
Nevertheless, the larger cycles may be needed to kick-start the discovery process of the smaller learning cycles. 

In this sense, one option to integrate human scientists and AI-driven science is for humans to build the context of the greater cycle (for example, fulfilling the role of the `Final Theory' and `Background knowledge' steps at the leftmost smaller cycle in Fig.~\ref{closeloop}), feeding the AI with new insights, and leave the AI to independently deal with the smaller cycles (such as the rightmost smaller cycle in Fig.~\ref{closeloop}), guided by the greater ones. The LLM's could, for example, be very useful as a technical interface and translation of human high-level larger cycle aspirations and their respective "divide-and-conquer" breakdown into smaller cycles. If one aims at the highest degree of automation of the discovery cycle, more sophisticated forms of AI should include automation of the validation, dissemination, refereeing, and other aspects of human science and its practice.

To tackle such challenges, we propose in the following sections the steps and technology suggested to conduct an entire cycle of AI-led scientific discovery~\cite{Gil2014-ch}, as in Fig.~\ref{closeloop}. 

\subsection*{\textit{Hypothesis Generation}}\label{sectionHypothesisgeneration}

One of the central components of the scientific practice is the `hypothetico-deductive' method \cite{Popper1972,King2011}. 
An additional set of epistemological tools is induction \cite{Russell1912}, abduction \cite{King2009} and counterfactual reasoning\cite{Pearl1995}.
To automate those knowledge processes, a deduction can be combined with simulation to infer the experimental consequences of hypotheses. Matching simulation with experimental output will be a reliable basis for an AI to accept or reject a hypothesis.
Such experimental output is tested with multiple interventions in the automated series of perturbation analyses \cite{Zenil2020cnat}. 
However, while one traditional approach to automate induction may follow, for example, new methods for clustering and regression, automating abduction and the creation of counterfactual scenarios may pose an even more challenging problem.
For this purpose, it would require the AI algorithm to explore irreducibly novel possibilities that are emergent to the current state of knowledge in which the AI is situated \cite{Abrahao2022}.

In this sense, neural networks are unlikely to be useful in the process of hypothesis generation, nor is any statistical machine learning. This is because they need training, and not only is training over hypothesis generation exactly the problem to be solved in the first place, but training over previous hypotheses, dividing them into rejected or valid, may undermine the freedom and the unbiased exploration that is desired of regions of interest in the hypothesis space. 
For hypothesis generation, what is needed is a bottom-up approach (e.g., a model-driven AI) or a hybrid one able to conduct cycles of systematic hypothesizing, 
from either partial or exhaustive enumerations (even if redundant though universal)~\cite{Morgan1971-ly, Thieme2005-ij}.

A bottom-up approach that deals with this open-endedness concerning the role of novelty is the field of algorithmic information dynamics (AID) \cite{Zenil2020cnat}, a framework for causal discovery and causal analysis based on algorithmic information theory and perturbation analysis.

Open-ended innovation in hypothesis generation and how to create and search over unbounded hypothesis spaces in less well-specified domains is an open challenge in itself, where research on the topics of this document can help make progress. These spaces and the methods exploring them usually have to deal with problems of intractability or uncomputability \cite{Zenil2018-pk, Zenil2019}.

Each method has its advantages and drawbacks and lies at different extremes of the causal inference spectrum.
Guiding heuristics based on first principles are needed to explore the hypothesis space~\cite{Lenat1982-aj}. Dovetailing partial results is necessary to avoid infinitely long cycles running the search. Here aspects of computability and tractability will be in play at every step, which we will need measures to deal with unless less powerful techniques are implemented (e.g. propositional logic or domain-restricted spaces such as a set of genetic circuits).
At one extreme are the statistical tools that confound correlation and causation but can help scientists make a call and guide their experiments, viz., graphical models that combine probability with symbolic logic, reasoning, and interventional calculus. 
The statistical approach often leads to less computationally expensive methods and, although in general, they may present distortions or biases toward some selected features \cite{Zenil2017a,Zenil2020}, it returns sound results in cases one knows a priori that the underlying generative processes are purely stochastic, stationary and ergodic.
At the other extreme is AID, which searches for sets of agnostic generative models compatible with observations, and exploits these models as testable underlying mechanisms and causal first principles~\cite{Zenil2018-pk, Zenil2019}, regardless of those being stochastic, computable, or mixed processes.
In addition to offering less constrained methods, for example, deconvolution algorithms \cite{Zenil2019b} and optimisation in non-differential spaces \cite{HernandezOrozco2021}, this approach offers results in direction to tackling the abduction and counterfactual problem, as for example shown in new methods for open-ended evolutionary computation \cite{Hernandez-Orozco2018,Hernandez-Orozco2018a}, and synergistic distributed computation \cite{Abrahao2017,Abrahao2018}.
However, bottom-up approaches like AID may not be humanly understandable, or when they are, scrutinising them  may require great computational effort, as is the case in other areas such as automatic theorem proving (e.g., the four-colour theorem).
LLMs may here again provide an advantage to interface between these model spaces as natural language processors integrating otherwise disparate systems translating among different domain databases and knowledge bases.

\subsection*{\textit{Experimentation and Sensing}}\label{sectionExperimentaionSensing}

One key task is to create AI systems for scientific discovery able to conduct experimentation and hypothesis testing independent of human instruction or with little to no human instruction. 
This is because what is desired to take scientific discovery to the next level is not the programming of algorithms able to conduct experiments, but open-ended algorithms able to set their own goals and experiments guided by previously conducted experiments (their own or from the human literature).
To this end, involving the machine embodiment to perform as a physical scientist by combining sensing and action together in the fully automated smaller cycles (which in turn are part of the larger encompassing AI-led closed-loop of scientific discovery) of empirical hypothesis testing, instrument-driven approaches render robotics key to making progress in physical experimentation so that more and more of the physical execution of experiments will be done using robotics \cite{Kitano1997}. 
This will increase the productivity of science, as robots work cheaper, faster, more accurately, and for longer than humans. 
Furthermore, if not embodied, the scientific experiment may collapse into a problem of data analysis and inference without the hypothesis, model, and theory testing that requires positive or negative feedback from the empirical side. Thus only a tiny part of the scientific discovery cycle would be tackled. 

Neural networks can help physical machines to embed themselves in a physical world for representation purposes, as neural networks have proven useful in representing all sorts of images. Still, innovation in areas of robotics and mechatronics will be required to accommodate the kind of depth and range of scientific experiments, in particular when it comes to accuracy and precision---which should not present a problem---while also helping with the current, very human problem of reproducibility~\cite{Baker2016-cd}. 
This is expected to have a significant impact on the reproducibility of science, as automating science requires semantic precision.
LLMs will also interface between human and robot instructions making it easier to create tools to automate experiments in natural language effectively instantiating a robot assistant able to process human instructions for scientific experimentation.

\subsection*{\textit{Rejection, Validation and Model Selection}}\label{sectionRjectionValidationModelselection}


Model selection and reduction have been a recurring theme across several sub-fields of areas, such as computational biology and neuroscience, with special reference to dynamical forward models. The idea is that if a complex nonlinear model can be reduced in complexity (fewer state variables and parameters), the investigator can more readily discern which parameters and state variables are more crucial to the model's behaviour, facilitating model analysis and understanding. One example is the reduction of the four-dimensional Hodgkin–Huxley model to a two-dimensional FitzHugh–Nagumo (FHN) system~\cite{Lindner1999-wy}. The core idea was to perform a time-scale separation into fast and slow subsystems. This has been used in several model reduction studies, including the cell cycle.
Techniques for dimension reduction, feature, and model selection will be helpful at this stage, from statistical approaches such as principal component analysis to more sophisticated ones such as minimal information loss techniques.

Another core idea for model selection is that each hypothesis formed will have a predicted probability of being correct,
possibly along with the associated cost of the respective experiment. This may be the monetary cost of executing the experiment, plus a temporal discount rate to value finding results more quickly. It has been empirically shown that using a Bayesian approach to experiment selection is sound and outperforms experiments chosen manually \cite{King2004a}. 

Current AI has shown the ability to yield valuable insights from noisy or incomplete data, optimise procedure design, and learn notions of structure amongst heterogeneous observations.  Neural networks have shown utility in isolating proper signals from noisy datasets spanning disciplines from physics to biology; such capabilities could be critical to establishing scientific conclusions as we reach the practical limit of experimental data quality~\cite{Drton2017-bz, Eddy2004-ub}.  Approaches from optimisation have demonstrated an ability to reduce the expense of experimental campaigns by optimising sampling patterns using, for instance, bandit-style methods to more rapidly design electric batteries or iteratively specify experimental conditions in biology.  Structure learning techniques from the graphical model literature could find use in identifying statistically meaningful relationships from large amounts of unannotated data~\cite{Drton2017-bz}.

\subsection*{\textit{Knowledge Representation and Natural Language Processing}}\label{sectionKnowledgerepresentationNaturallanguageprocessing}

Ingested knowledge may no longer be machine-readable, either rule-based or probabilistic given that LLMs can interface between them but its possible caveats, such as low-level hidden misalignments, are difficult to unveil, making difficult traceability and liability. LLMs can allow machines to read, interpret, and exploit the current knowledge from a scientific domain in human natural language and digest the relevant literature in the target area. 
An AI-led scientific discovery approach will require at least access to the space of interest needed for the system to be able to validate or reject a hypothesis based on contradiction or confirmation of previous knowledge which may be difficult in a black box like an LLM. So, the LLM will need to be self-explanatory with the caveat that the output explanation may not fit the internal statistical derivation of what the LLM ends up producing. An independent system and a more explainable mechanistic process may need to verify the output.
Without LLMs, this task would have required massive databases and curation efforts for domains that are not already significantly represented in a computable fashion.
Although all sorts of languages can be used to represent knowledge, some domains will be aptly represented by propositional-logic rules, such as simplified genetic circuits, to avoid these potential misalignments from LLMs or statistical ML in general.
Other domains will require more sophisticated representations, either to encompass the greater complexity of an extended domain or to deal with the greater sophistication of, e.g., a domain such as biomedicine, where system-expert rules with ifs, dos, and whiles are required, hence the full power of first-order logic and Turing-completeness. 

For example, knowledge representation systems/ontologies are well developed in biology: The Gene Ontology (GO), nascent Causal Activity Models with the GO, Human Phenotype Ontology, Chemical Entities of Biological Interest, Ontology of Biomedical Investigation, among others~\cite{Stevens2000-fu, Bard2004-ej}. So are integration efforts built on these ontologies, e.g., Monarch~\cite{Shefchek2020-ci}.
The JST MIRAI `Robotic Biology' project can also provide technologies to help adoption, such as LabCode, a common formal language for experimental protocols, LabLive, a laboratory information IoT platform, and real-time parallel workflow scheduling software that can decompose processes in a given protocol and assign each to different robots/equipment so these are executed considering dependencies and concurrency between them.

Another example is statistical relational learning (SRL), which combines relational learning and probability theory and is an area of ML research (e.g. \cite{Raedt2008}),  
enabling the representation of beliefs about relational data using probabilistic models. 
Relational Learning (RL) is a general representation language based on first-order predicate logic \cite{Raedt2008}.
Such probabilistic logic models enable the specification of graphical models (Bayesian networks, Markov networks, etc.) over large relational domains. 
One of the fundamental design goals of the representation formalisms developed in SRL is to abstract away from concrete entities and to represent instead general principles that are intended to be universally applicable. A key advantage of RL is that it can easily incorporate background scientific knowledge, and learn about structured objects such as scientific models particularly appropriate for utilising background bioinformatic data \cite{Orhobor2020}.
These approaches can be further enhanced or complemented by the do-calculus \cite{Pearl1995,Pearl2012} or algorithmic information dynamics \cite{Zenil2020cnat}.

Deep neural networks are also good at capturing the apparent granularity and complexity of natural phenomena in a computable form (in weighted vectors of numerical matrices). The success of neural networks implies that once one captures an object in an optimal way, classification is trivial, as it was for deep learning in the protein-folding challenge~\cite{Wei2019-vt, Jumper2021-gb} with its limitations.

Assuming that an appropriate formalism to record observation could be found for any domain, a modeller may be faced with a severe feature selection problem, which translates into a question of the identity of the relevant state variables of the systems of interest, e.g., drug docking dynamics for drug discovery or cytokines for cell dynamics. 
On the one hand, all the system entities that are measured could define the set of state variables to be represented, e.g. drugs or proteins, augmented with the set of rules to which the entities may be subjected, such as thermodynamics or collisions. 
However, this type of representation could quickly become very complex~\cite{Tang2019-ni}. 
On the other hand, a certain subset of combinations of measured state variables may be a useful representation of the governing dynamics driving a possible system, and this is a question that needs to be asked and resolved for scientific domains on a case-by-case basis.
Such a feature selection problem in computably representable objects is often found in analyses in which one is assuming a pure stochastic nature of the system's generative processes, although the system also comprises deterministic, mechanistic, or computable subprocesses \cite{Zenil2017a}. 
In addition, even in cases the whole algorithmic space of possibilities is covered, analyzing the information content carried by a network highly depends on the multidimensional space into which it is embedded \cite{Abrahao2021}, where distortions may be exponential for multidimensionality-agnostic encodings.

Thus, developing expressive and efficient frameworks to computationally represent and capture a wide range of scientific knowledge about processes, models, observations and hypotheses is key. 
Additionally, in the opposite direction of knowledge representation by machines, the AI for scientific discovery may need to communicate in the form of a publication or other scientific means to explain the innovation and methods behind the discovery to humans and to articulate its significance and impact.
Thus, not only we will have to improve knowledge representation \cite{Evans2011} of these scientific objects of inquiry, but also include (meta)knowledge representation of the social dynamics constituted by the scientific practice of humans and AI algorithms in the loop.
This in turn should lead to better mitigations of the aforementioned problems of reproducibility and biases in science.
Capturing scientific knowledge will push the limits of the state of the art.

A choice that has to be made, on a case-by-case basis, is whether it is required that AI conducts the experiments without much human understanding or whether it is acceptable not to have a sophisticated translation of both the hypotheses generated and the process arriving at a conclusion. 
In cases where there is a requirement for human understanding, and even for the most general case, at least partial interpretation by human scientists may be required. 

Thus, knowledge representation and natural language processing techniques will be needed to be jointly developed to both: 
feed the system with the current knowledge relevant to the hypothesis space;
and guide the search (in cases of human-machine interaction) or be able to follow up the inference process and interpret the results~\cite{Chowdhury2005-gl,Cambria2014-yd}. 
These requirements will force us to make progress on humanly readable and interpretable machine-human translation.

\subsection*{\textit{Integration, Interpretation and Interfacing}}\label{sectionIntegrationInterpretationInterfacing}

One of the most challenging aspects of scientific discovery is integrating a new piece of information with the corpus of existing human knowledge.
Analysing the data will require moving to the larger learning loop where there is a broader view of the results for possible (re-)interpretation. 
This is because while the specific objective for the target hypothesis may have been rejected, one of the main serendipity checkpoints is the reinterpretation of results in a broader context. 

Machine learning systems have proven incredibly useful for automated knowledge base construction. They have recently contributed to creating multiple large databases describing, for instance, genome-wide association studies and drug-disease interactions directly from the published literature~\cite{Andronis2011-uk}.  This ability to create massive knowledge bases that rapidly and effectively contextualise new findings could substantially accelerate scientific discovery by ensuring that seemingly disparate dots are more rapidly connected.

However, exploring and understanding user context requires automating certain social, political, and economic aspects of interconnected knowledge that are intrinsic to science \cite{Fortunato2018}.
The AI systems' interactions with scientists must be guided by a knowledge-rich multi-agent model \cite{Kitano1997} that enables the AI systems to act as colleagues like LLMs now may allow. 

This constitutes an inextricable loop in which human scientists and AI-scientists are parts of a whole system, which the AI algorithm should try to optimise.
A striking example of such an optimal interplay has been the evolution of machine-human chess collaboration. 
After the defeat of Gary Kasparov, it became standard to have human chess players practice with computers, and for champions, it became impossible to reach the level of playing demanded without intensive computer training~\cite{Campbell2002-tv}. To this day, the strongest freestyle chess teams have been those able to strike a perfect balance between machine and computer training and playing. 

Again, neural networks and statistical machine learning will not help in this process, at least not on their own or in their traditional architectures. 
What is most likely needed here is first an inference engine able to extract knowledge readable by humans as well, especially under human-machine schemes. 
Classical logical inference engines are key, but so are hybrid approaches combining statistical learning and symbolic computation so that the AI algorithms' objectives and their respective performance measures are not always fixed in advance \cite{McCarthy2007-ns}.
Techniques such as feature selection and data dimension reduction will be helpful in this regard.
Secondly, an AI algorithm that can simulate the network topological properties of scientific production \cite{BarabSci} and perform the steps of the full cycle of AI-led scientific discovery, while taking into account the relational structures and biases that emerge when the AI-human relationship is analysed as a single system.

The application of AI to science will confer multiple advantages, and eliminate some of the disadvantages of having a human in the loop, such as biases and lack of reproducibility. Yet, if humans rely on automated scientific discovery, verifiability and transparency are crucial because the coupled AI-human system has to be able to be formally verified to ensure that it matches the goals and that the results match the process.
In this manner, the AI algorithm should be designed to continuously reiterate its data gathering from the outputs and behaviours of the whole system the AI is part of.
The same for the human scientist, which needs to be able to perform, evaluate, and produce analytical reasoning while participating in this coupled computational-social system.
This in turn may give rise to innovative methodologies and epistemological grounds that foster the scientific justification of the results and novelties discovered by such a coupled system.

\subsection*{\textit{Closing the Loop}}

Finally, connecting all the steps will require a meta-algorithm that will need to systematically manage each cycle and even decide when to break or restart the cycles (see Fig. \ref{closeloop}), if human intervention is taking place.
The whole cycle should be open to human intervention, and the AI algorithm should both reiterate the new insights and data given by humans and counter any bias that these may introduce.

Technology for remote web control and monitoring of full-cycle scientific discovery may require technologies such as TypeScript, React, GraphQL, Jest, and Redux to create a web-based beamline control system.
Techniques such as optimisation and anomaly detection can be used to find possible gaps and even glitches (found or promoted). These gaps can be exploited to reinterpret data, explore other regions of the hypothesis space and kick-start the process of hypothesis generation again, thus closing and restarting the discovery cycle.

Notice that each of the above aspects of the AI-led closed-loop science can be considered landmark projects that will also require solutions to many standard technical problems \cite{Kitano1998-eo}.
Therefore, toward being able to close the loop with an AI-led science, the ``grand challenge'' \cite{Kitano1998-eo} that we propose ranges over automating not only laboratory practices and theory making, but also writing a paper, refereeing, and disseminating achievements.

\section*{Conclusion: the Future of AI in Scientific Discovery}\label{sectionConclusion}

Future scientific progress has become almost unthinkable without the involvement of machine learning. We have explored some challenges and opportunities in utilising and exploiting AI. We argue that a closed-loop formulation not only augments and accelerates scientific discovery but also leads science in new directions, thus potentially disrupting the future trajectory of human science. Such closed-loop experimentation led by AI may also mitigate current challenges, such as the production and replication of data. 

The development of AI to discover new fundamental scientific laws and representations is different compared to the application of AI to games such as chess\index{chess}, shogi\index{shogi}, or Go\index{Go}. However, recent developments surprisingly suggest that some scientific challenges may not be that different from these games~\cite{Silver2016,Hassabis2017,Kitano2021}. 

New questions for scientists and policymakers are increasingly pertinent. For example, do we require AI equipped with sufficient intelligence and autonomy to render it capable of sensing and making observations to ask novel scientific questions? Who should control AI4Science systems, humans or other tertiary systems we may trust? How will the role of the future scientist change? Yet, these challenges must be solved since we urgently need to solve problems like cancer and climate change.

\bibliographystyle{Science}

\end{document}